\documentclass[11pt]{article}

\makeatletter
\def\blfootnote{\gdef\@thefnmark{}\@footnotetext}
\makeatother

\usepackage[preprint]{acl}

\usepackage{times}
\usepackage{latexsym}
\usepackage{hyperref}

\usepackage{float}

\usepackage{tcolorbox}
\usepackage{graphicx}
\usepackage{subcaption}
\usepackage{booktabs}
\usepackage{latexsym}
\usepackage{enumitem}
\hypersetup{
    colorlinks=true,
    linkcolor=black,
    citecolor=blue,
    urlcolor=blue,
    filecolor=magenta,
    linkbordercolor={1 1 1},
    pdfborder={0 0 0}
}

\usepackage[T1]{fontenc}

\usepackage[utf8]{inputenc}

\usepackage{microtype}

\usepackage{inconsolata}

\usepackage{graphicx}

\title{Patterns vs. Patients: Evaluating LLMs against Mental Health Professionals on Personality Disorder Diagnosis through First-Person Narratives}

\author{
  \textbf{Karolina Drożdż\textsuperscript{1*}},
  \textbf{Kacper Dudzic\textsuperscript{1,2,3*}},
  \textbf{Anna Sterna\textsuperscript{4}},
  \textbf{Marcin Moskalewicz\textsuperscript{1,4,5}}
\\
\\
  \textsuperscript{1}IDEAS Research Institute, Warsaw, Poland \\
  \textsuperscript{2}Adam Mickiewicz University, Poznań, Poland \\
  \textsuperscript{3}AMU Center for Artificial Intelligence, Poznań, Poland \\
  \textsuperscript{4}Poznań University of Medical Sciences, Poznań, Poland \\
  \textsuperscript{5}Maria Curie-Skłodowska University, Lublin, Poland \\
\\
  \small{
    \textbf{Correspondence:} \href{mailto:karolina.drozdz@ideas.edu.pl}{karolina.drozdz@ideas.edu.pl}
  }
}

\tcbuselibrary{breakable}
\begin{document}
\maketitle
\begin{abstract} Growing reliance on LLMs for psychiatric self-assessment raises questions about their ability to interpret qualitative patient narratives. This depth over breadth case study directly compares state-of-the-art LLMs and mental health professionals in assessing Borderline (BPD) and Narcissistic (NPD) Personality Disorders based on Polish-language first-person autobiographical accounts. Within our sample, the overall diagnostic scores of the top-performing Gemini Pro models (65.48\%) were 21.91 percentage points higher than the average scores of the human professionals (43.57\%). While both models and human experts excelled at identifying BPD ($F_1 = 83.4$ \& $F_1 = 80.0$, respectively), models severely underdiagnosed NPD ($F_1 = 6.7$ vs. $50.0$), showing a potential reluctance toward the value-laden term ``narcissism.'' Qualitatively, models provided confident, elaborate justifications focused on patterns and formal categories, while human experts remained concise and cautious, emphasizing the patients' sense of self and temporal experience. Our findings demonstrate that while LLMs might be competent at interpreting complex first-person clinical data, their outputs still carry critical reliability and bias issues.
\end{abstract}

\section{Introduction}

Systemic pressures on mental healthcare have led to critical access gaps and diagnostic delays~\cite{Sun2023, Barbui2025}. Consequently, the public is increasingly turning to widely available Large Language Models (LLMs) for self-assessment, thereby bypassing traditional clinical pathways~\cite{Lawrence2024, McBain2025}. While this trend may partially relieve strain on the healthcare system, it also introduces considerable ethical and clinical risks, such as the potential for inaccurate or misleading diagnoses~\cite{Harrer2023}. This study empirically compares the diagnostic capabilities of current open- and closed-source LLMs to mental health professionals in identifying a complex and frequently misdiagnosed condition of Personality Disorder (PD). 

Previous research shows that state-of-the-art LLMs can achieve performance levels that approach, or in some cases surpass, those of trained clinicians on standardized medical benchmarks~\cite{grzybowski-etal-2025-polish, workum2025}, and substantially improve clinicians’ diagnostic accuracy when used as a decision-support tool~\cite{noda2025}. These capabilities extend to chatbot-patient interaction, where recent investigations suggest that LLMs can surpass human physicians in expressing empathy in written clinical communication~\cite{ayers2023}. However, it remains unclear whether such capabilities can be effectively applied to complex psychiatric conditions, where diagnoses often lack well-established objective markers and depend heavily on nuanced, subjective accounts of patients. 

A limited body of work evaluating LLMs specifically on psychiatric and psychological benchmarks reveals a more uneven performance. Studies show that models such as GPT-4 and Llama 3 achieve high accuracy (up to 85\%) on binary mental disorder detection tasks, but their performance varies significantly across datasets~\cite{hanafi2025comprehensiveevaluationlargelanguage}. Notably, LLMs fine-tuned on domain-specific data do not outperform generalist models on existing benchmarks~\cite{fouda2025psychiatrybenchmultitaskbenchmarkllms, nguyen-etal-2025-large}.

Existing evaluations typically rely on simplified formats and noisy or synthetic datasets, which are not comparable to the complex, first-person nature of psychiatric patients' testimonies. Furthermore, by prioritizing outcome metrics over explanatory reasoning, current studies fail to verify whether models emulate human cognitive processes or merely guess accurately. To address this gap, our study introduces a methodological shift from breadth to depth. Specifically, instead of evaluating large volumes of surface-level material, we deliberately focus on
a smaller number of raw, in-depth clinical cases. We apply this depth over breadth approach to Borderline Personality Disorder (BPD) and Narcissistic Personality Disorder (NPD). These are strong, yet underexplored, phenomena for model evaluation, particularly as their diagnostic assessment might be overdetermined by the theoretical framework applied (or the lack thereof, as with the lay understanding of the terms ``borderline'' and ``narcissism'').

Recent revisions in DSM-5~\cite{APA2022} and ICD-11~\cite{WHO2022} have redirected attention from a distinct \textit{categorical} approach to PD to one based on transdiagnostic commonalities and their continuous variation, i.e., a \textit{dimensional}, non-exclusive approach. Given the prominence of the ICD and DSM frameworks, current LLMs could be biased toward a dimensional understanding of PD. Conversely, the ubiquity of terms like ``borderline'' and ``narcissism'' in everyday language---and thus in training corpora---suggests a potential counter-bias toward a categorical understanding of PD. Since empirical evidence does not clarify whether current LLMs lean toward a dimensional or categorical view of PD, we examined both frameworks independently. 
\looseness=-1

\enlargethispage{\baselineskip}

In summary, our contributions are as follows:
\begin{itemize}[nosep]
    \item We present the first evaluation of state-of-the-art LLMs on raw, first-person accounts of patients' life stories, assessing the models' capacity to interpret lived biographical experiences.
    \item We introduce an expert-informed evaluation protocol that encompasses diverse conceptualizations of mental disorders (i.e., categorical vs. dimensional). 
    \item We directly compare the diagnostic performance of human experts and LLMs on complex mental disorders, moving beyond isolated model benchmarking.
    \item We prioritize depth over breadth, adopting an explanatory approach to both human data and model outputs, aiming not only to quantify performance but also to provide interpretable insights.
\end{itemize}

\section{Methods}
\subsection{Data}
The database comprises six clinical cases---three BPD and three NPD---alongside one healthy control. These were selected from a larger dataset of first-person symptomatology, which includes 24 BPD cases, 20 NPD cases, and 20 healthy controls (HC), classified according to ICD-10 diagnostic criteria.

To ensure the clinical validity of the sensitive first-person data used for evaluation, clinical narratives were collected at the Psychiatric Hospital in Międzyrzecz, Poland, following a multi-stage procedure. Participants enrolled in the original study were diagnosed as outpatients, then confirmed as inpatients; each diagnosis was confirmed by four independent clinical experts: two psychiatrists and two psychologists. To ensure comparability, categorical diagnosis was accompanied by personality functioning and maladaptive traits assessment, using a multi-dimensional protocol (see \autoref{app:instruments}). HC participants were healthy volunteers who responded to the study’s announcement and declared no diagnosed psychiatric conditions or felt symptoms. They were subsequently screened for personality pathology using LPFS BF 2.0 ~\cite{Lakuta2023}  and PICD~\cite{Cieciuch2022}.

All the participants took part in 50--70 minute semi-structured qualitative interviews adapted from McAdams’ Life Story Interview~\cite{McAdams1988}, designed to explore narrative identity. Two additional questions probing reflective self-experience were included. Interviews were audio-recorded and transcribed, resulting in a Polish-language corpus exceeding 200,000 words. 

To extract cases for this study, we followed intensity sampling to select information-rich, yet non-extreme cases representing: integrated personality as defined in DSM-5 (HC, $N=1$) and mild, moderate, and severe levels of personality impairment (clinical sample: BPD, $N=3$; NPD, $N=3$). The severity variance represented in the sample was meant to capture the breadth of narrative expression within both BPD and NPD. 

The final sample size was set to balance the necessity to represent the heterogeneity of PD symptomatology with the practical constraints imposed by a human evaluation. The final database, amounting to 27,641 words, was evaluated for approximately five hours per case on average, thus reaching the maximum volume that avoided overwhelming human evaluators and secured the reliability of their assessments.

\subsection{Model Selection}
We included a total of ${N=16}$ leading models and variants, split into a closed-source ($N=9$) and an open-source group ($N=7$). Our selection criteria encompassed: public interest and availability, presence in recent evaluations on related material~\cite{fouda2025psychiatrybenchmultitaskbenchmarkllms, hua2025scoping, nguyen-etal-2025-large}, access model (licensing), parameter count, country of origin, and reasoning ability. 

The closed-source group included: Gemini 2.5 Pro~\cite{comanici2025gemini25pushingfrontier}, Gemini 3 Pro~\cite{google2025gemini3}, Claude Opus 4.1~\cite{anthropic2025claude4}, GPT-4o~\cite{openai2024gpt4ocard}, GPT-4.1~\cite{openai2025gpt41}, and four variants of GPT-5~\cite{openai2025gpt5} with all available values of the OpenAI API's reasoning effort parameter (from \textit{minimal} to \textit{high}). The open-source group consisted of: Gemma 3 27B~\cite{gemmateam2025gemma3technicalreport}, Llama 3.3 70B~\cite{grattafiori2024llama3herdmodels}, DeepSeek R1 0528~\cite{deepseekai2025deepseekr1incentivizingreasoningcapability}, two variants of DeepSeek v3.1 Terminus~\cite{deepseekai2025deepseekv3technicalreport} with reasoning either enabled or disabled, and two variants of Qwen 3 32B~\cite{yang2025qwen3technicalreport}, also with and without reasoning. 

We did not include domain-specific models trained on medical data for two reasons. Firstly, the context windows of available medical models evaluated in existing literature~\cite{fouda2025psychiatrybenchmultitaskbenchmarkllms, hanafi2025comprehensiveevaluationlargelanguage, hua2025scoping, nguyen-etal-2025-large} proved to be too small for the full patient testimonies---except for MentalQLM~\cite{shi2025mentalqlm}, which has not been made publicly available. Secondly, recent studies indicate that general-purpose models perform similarly to or better than medical models on psychiatric tasks~\cite{fouda2025psychiatrybenchmultitaskbenchmarkllms}.

\subsection{Mental Health Experts}
The study sample consisted of $N=6$ highly experienced mental health professionals, comprising three psychiatrists and three psychotherapists, recruited through our professional network. They received no compensation for their participation in the study. The group was balanced by gender, with three participants identifying as men and three as women; the mean age was 48.5 years (SD = 11.18), with an average of 18.5 years of professional experience. Importantly, the experts were external to the original diagnostic teams. They were blinded to the patients' previous medical records and ground truth diagnoses, and had never previously encountered the patients whose narratives they evaluated. Experts formulated their diagnostic decisions independently of one another.

\subsection{Procedure}
Human and model participants conducted a diagnostic assessment of each autobiographical testimony, adhering to a 6-step response template (see \autoref{app:model_prompt}). The protocol required participants to assign a (1) categorical diagnosis and (2) severity rating, each with a corresponding confidence level (3 \& 4). Participants were also instructed to provide a brief qualitative justification for both (5) the diagnosis and (6) severity assessment, in no more than 100 words, indicating relevant evidence from the testimony and outlining their theoretical understanding of PD and their origins. All assessments were conducted in Polish, reflecting the language of the source data and the native language of the mental health professionals.

To avoid priming and to allow participants to apply their expert knowledge, the categorical diagnosis was an open-ended task without predefined options. In contrast, both the severity rating and the confidence ratings followed fixed scales: severity was rated on a 0--3 scale (0 = none, 1 = mild, 2 = moderate, 3 = severe), and confidence on a 1--4 scale (1 = guessing, 2 = somewhat confident, 3 = fairly confident, 4 = completely confident). 

Importantly, to reliably evaluate the baseline competence of the models, no model-only in-context learning paradigm was employed in the protocol. This is because potential differences in understanding the patients' data between human participants and models were another point of interest of the evaluation. The differences could potentially stem from, e.g., a lack of such priming through example human responses.

While human participants evaluated each testimony once, models were presented with each case three times to address potential diagnostic inconsistency stemming from their non-deterministic nature. This resulted in 7 trials per human expert and 21 trials (7 cases \texttimes~3 repetitions) per model.

\subsection{Data Analysis} \label{sec:data-analysis} 
\subsubsection{Performance Metrics}
Two complementary performance metrics were computed: a categorical and a dimensional score. The first, based on the open-ended diagnosis,  captured performance within the traditional categorical model, which assumes the presence or absence of discrete diagnostic entities. The second was calculated from the severity rating, which reflects the dimensional model conceptualizing personality disorder along a continuum of severity. 

For human participants, each score represented the total number of correct evaluations across the seven cases (0--7). For the models, scores were computed across all 21 trials to account for reliability (7 cases \texttimes~3 repetitions; 0--21 score). Since a correct model response required both accuracy and consistency across the repeated trials, a perfect score would indicate not only a valid diagnostic judgment but also a fully consistent performance.

\subsubsection{Diagnostic Justifications}
Human mental health professionals can arrive at similar diagnostic conclusions through different cognitive and emotional pathways \cite{biondi2022clinician}. Traces of these mental processes are reflected in the semantic content of the justifications provided during the diagnosis. To use this information alongside quantitative metrics, we mapped the diagnostic justifications onto a high-dimensional semantic embedding space. This allowed for a comparison of diagnostic expertise leverage between human and model participants, as well as among different models.

\texttt{BAAI/bge-multilingual-gemma2}\footnote{\url{https://huggingface.co/BAAI/bge-multilingual-gemma2}} was chosen as the embedding model owing to its superior performance on Polish-language tasks in the MMTEB benchmark~\cite{enevoldsen2025mmtebmassivemultilingualtext}. We created a single summary embedding representing the averaged semantic content of justifications for each model separately, as well as a single one representing all human participant data to account for its comparative scarcity. A more detailed process of embedding creation was described in \autoref{app:justification_process}. The summary embeddings were aggregated into two distinct datasets: a global dataset comprising all agents (both models and humans) to assess the human-AI semantic gap, and a model-only dataset to allow for a finer-grained analysis of inter-model semantic differences. Given the embeddings of each justification type, a pairwise cosine similarity matrix was calculated to enable a dimensionality-reduced visualization of the similarities.

Due to the depth over breadth approach of the study, the summary embeddings constituted an unusually small set of 17 data points, making a faithful visualization of the high-dimensional relationships between these representations harder. Accordingly, we employed Multidimensional Scaling (MDS)~\cite{kruskal1964multidimensional} as the dimensionality reduction algorithm through the \texttt{scikit-learn}\footnote{\url{https://github.com/scikit-learn/scikit-learn}} library~\cite{JMLR:v12:pedregosa11a}. Unlike local manifold techniques---unreliable for small sample sizes---or linear projections---which prioritize variance over pairwise similarity---MDS was selected for its ability to preserve global metric distances, offering a deterministic representation that better reflects the relative semantic similarities between the summary embeddings than the alternatives would. Details on the hyperparameter values used can be found in \autoref{app:hyperparameters}.

\subsubsection{Lexical Features} 
To further investigate the divergence between linguistic justification of diagnostic reasoning in humans and models, we conducted a follow-up inquiry into interpretable differences. We aggregated all justifications into two distinct corpora---human-written and model-written---and sought to identify lexical features that were statistically overrepresented in one group relative to the other. The features were identified with the weighted log-odds ratio with the informative Dirichlet prior method originally applied to the problem of detecting lexical polarization in political discourse~\cite{monroe2008fightin}. The method is well-suited for analyzing small and/or imbalanced text corpora; it estimates the z-score of the log-odds ratio for each n-gram, controlling for variance in word frequency, which specifically addresses the issue of sample size imbalance between the human and model datasets in our use case.

In the technical implementation, we first applied a text pre-processing pipeline to the two corpora, converting non-Polish non-ASCII characters into ASCII equivalents, as well as removing LLM-characteristic text formatting artifacts, punctuation, digits, and optionally stop words. Finally, we applied Monroe et al.'s method through the ConvoKit library~\cite{chang-etal-2020-convokit}. We considered two parameters to adjust in the feature generation process: stop word removal---either enabled or disabled, and the maximum n-gram length---exclusive ranges of 1 to 3 and a combined 1--3 one, ultimately choosing unigrams with the stop words removed as the most informative configuration. 

\section{Results}
\subsection{Diagnostic Performance}

\begin{figure}[h]
    \centering
    \includegraphics[width=\columnwidth]{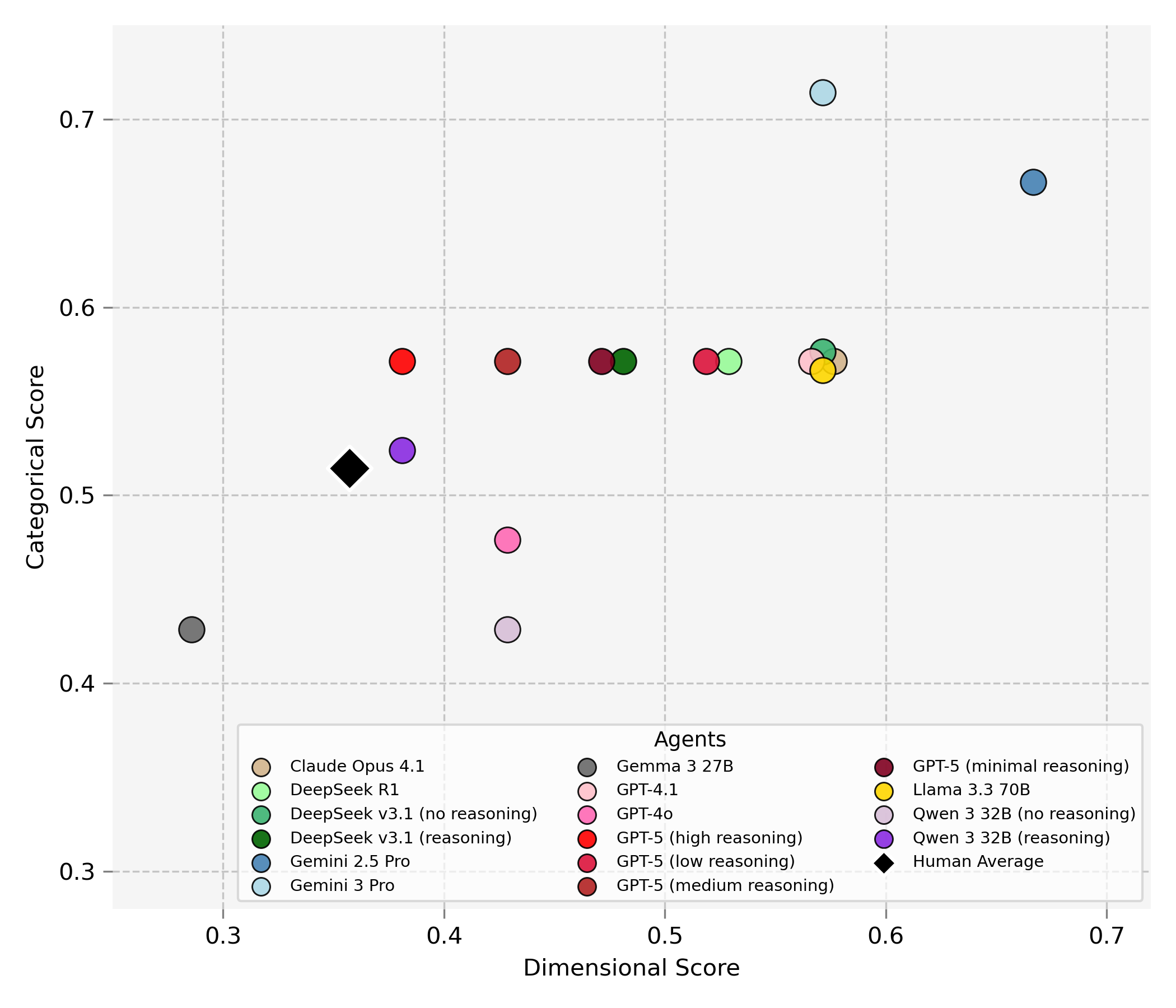} 
    \caption{Categorical and dimensional score (recalculated on a 0--1 scale) comparison between mental health professionals and models.}
    \label{fig:performance_comparison}
\end{figure}

\begin{figure}[h]
    \centering
    \includegraphics[width=\columnwidth]{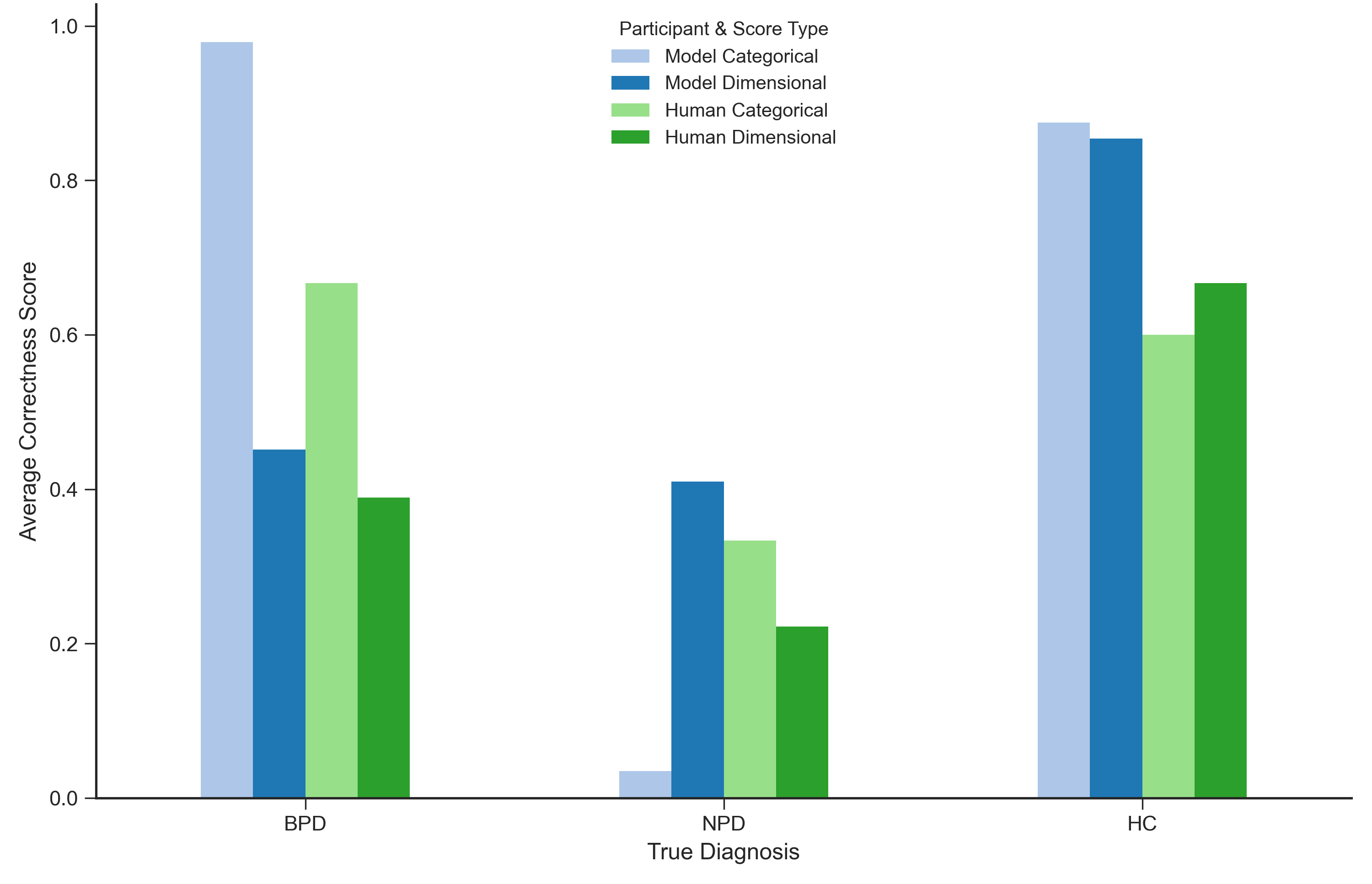}
    \caption{Average model and mental health professionals' scores by true diagnosis.}
    \label{fig:accuracy_dissociation}
\end{figure}

A comparative analysis of diagnostic performance revealed significant variability among models and mental health professionals, as summarized below.

\textbf{Top-performing models achieve a higher diagnostic average than human experts.} The overall human average score was 43.57\%, with a categorical score of 3.60 out of 7 (51.43\%) and a dimensional score of 2.50 out of 7 (35.71\%). An exploratory analysis of the participant sub-groups suggested a potential performance gap: psychotherapists scored higher on the categorical task (mean difference: 2.33), while psychiatrists scored slightly higher on the dimensional task (mean difference: 0.33). Importantly, inter-rater reliability among human experts was relatively low, reflecting the inherent subjectivity of psychiatric evaluation \cite{diforti2025interrater} restricted to the text modality \cite{fiquer2013talking}. For categorical diagnosis, Fleiss' Kappa was 0.11, and Krippendorff's Alpha (nominal) was 0.14, indicating slight agreement; for dimensional severity ratings, Krippendorff's Alpha (ordinal) was 0.22, indicating fair agreement.

Model performance, evaluated over 21 trials, was relatively dispersed, with several top-performing models exceeding the human average (see \autoref{fig:performance_comparison}). For the categorical task, most models (12 of 16) converged on a score of 12/21 (57.1\%), outperforming the human average of 51.4\%. Gemini 3 Pro achieved the highest categorical accuracy (15/21; 71.4\%), while Gemini 2.5 Pro performed best on the dimensional score (14/21; 66.7\%). Notably, the averaged Gemini Pro model family overall score exceeded the human overall average by 21.91 percentage points (65.48\% vs. 43.57\%). Conversely, Gemma 3 27B was the worst-performing model, scoring 9/21 (42.8\%) on the categorical task and 6/21 (28.5\%) on the dimensional task.

Further analysis of model performance indicated that reasoning did not consistently improve diagnostic accuracy. Additionally, no clear performance gap was observed between closed-source and open-source models. In contrast, model size appeared to be a factor, with smaller models generally performing worse than larger ones. However, drawing definitive statements about its influence is difficult, as they would rely exclusively on the open-sourced models in our sample. This is further compounded by the fact that model performance being correlated with model size is (mostly) a general principle of machine learning.

\textbf{Diagnostic distributions reveal frequent BPD overdiagnosis and NPD underdiagnosis tendencies, particularly among models.} An analysis of the $N=378$ total diagnoses (see \autoref{sec:appendix_table}, \autoref{tab:diagnoses_distribution}) by both models ($N=336$) and humans ($N=42$) revealed systematic diagnostic bias. 

BPD was the most common diagnosis, accounting for 53.97\% of all combined diagnoses. This tendency to overdiagnose BPD resulted in a near-perfect categorical recall for models (97.9\%, $F1=83.4$), substantially higher than that of human experts (66.7\%, $F1=80.0$). However, such sensitivity came at the cost of precision (72.7\%), whereas human experts demonstrated perfect precision (100\%), suggesting that while clinicians missed some cases, their positive diagnoses were fully reliable. The high categorical recall for both models and humans was often paired with a lower dimensional recall (45.14\% for models, 38.89\% for humans; see \autoref{fig:accuracy_dissociation}).

In contrast, NPD was severely underdiagnosed by both humans and models, representing 2.65\% (10 of 378) of all diagnoses, with humans outperforming the models (14\% against 1.5\% for models). Gemini 2.5 Pro and Gemini 3 Pro were the only models to correctly identify 1 out of 3 NPD cases (in 2 out of 3 trials for Gemini 2.5 Pro, and 3 out of 3 trials for Gemini 3 Pro). Consequently, the aggregate model recall for NPD collapsed to just 3.5\% ($F1=6.7$). While the rare NPD diagnoses provided by models were accurate ($precision = 100\%$), the vast majority of cases were missed. This contrasts with human experts, who achieved a higher recall of 33.3\% ($F1 = 50.0$). However, a distinct positive dissociation was observed in model performance: despite failing to categorically identify the disorder, models achieved a significantly higher dimensional recall (40.97\%) for the same NPD cases, notably outperforming human experts (22.22\%) on the severity metric.

\enlargethispage{\baselineskip}

The ``healthy'' label was the second most common, accounting for 21.43\% of all assigned diagnoses. Models were highly effective at identifying the absence of pathology ($recall = 87.5\%$, $F1=69.4$) compared to human experts ($recall = 60.0\%$, $F1=50.0$). Despite these high scores, a detailed analysis of diagnostic distributions (see \autoref{sec:appendix_d}, \autoref{fig:heatmap}) reveals a specific ``depathologizing bias'' in the GPT family. GPT-4o showed the most extreme bias, misclassifying cases as healthy 10 times. GPT-4.1 followed with 5 false positives, whereas the GPT-5 variants had 2--3 each. This suggests a tendency in the GPT model family to favor non-clinical classifications. Notably, humans also demonstrated a slight tendency toward non-clinical classifications, evidenced by 4 false positives (see \autoref{sec:appendix_d}, \autoref{fig:heatmap}), driven by frequent assessments that the testimonies lacked sufficient evidence for a PD diagnosis. \\
\indent A significant misclassification bias was observed for Avoidant Personality Disorder (AvPD) (see \autoref{sec:appendix_table}, \autoref{tab:diagnoses_distribution}). It emerged as the third most common diagnosis overall (15.08\%), but this prevalence was driven largely by models. AvPD accounted for 16.07\% (54 of 336) of model-generated diagnoses, compared to 7.14\% (3 of 42) of human diagnoses. Given that AvPD was not present in the ground truth answers, this high prevalence points to a tendency to misinterpret symptoms of other conditions as AvPD. On the other hand, only one out of 336 diagnoses by models indicated a PD category not present in current diagnostic frameworks, namely Masochistic PD. \\
\indent\textbf{Models demonstrate higher and more uniform confidence than human experts.} Overall, models displayed higher certainty than human practitioners for both categorical diagnosis ($M=2.94$ vs. $M=2.52$) and severity assessment ($M=3.07$ vs. $M=2.86$). Across both groups, certainty was higher for severity assessments than for categorical diagnoses. Among the individual models, Gemini 3 Pro ($M=3.69$) and Gemini 2.5 Pro ($M=3.40$), the top-performing models in this study, exhibited the highest levels of certainty. Conversely, GPT-5 with high reasoning effort was the least certain model ($M=2.57$). The most striking difference between groups was that the models never utilized the lowest certainty score (1 -- ``guessing'') whereas human practitioners used it in 19\% of diagnostic ratings and 11.1\% of severity ratings.

\subsection{Diagnostic Justifications}
\subsubsection{Semantic Embeddings}

\begin{figure}[h]
    \centering
    \includegraphics[width=\columnwidth]{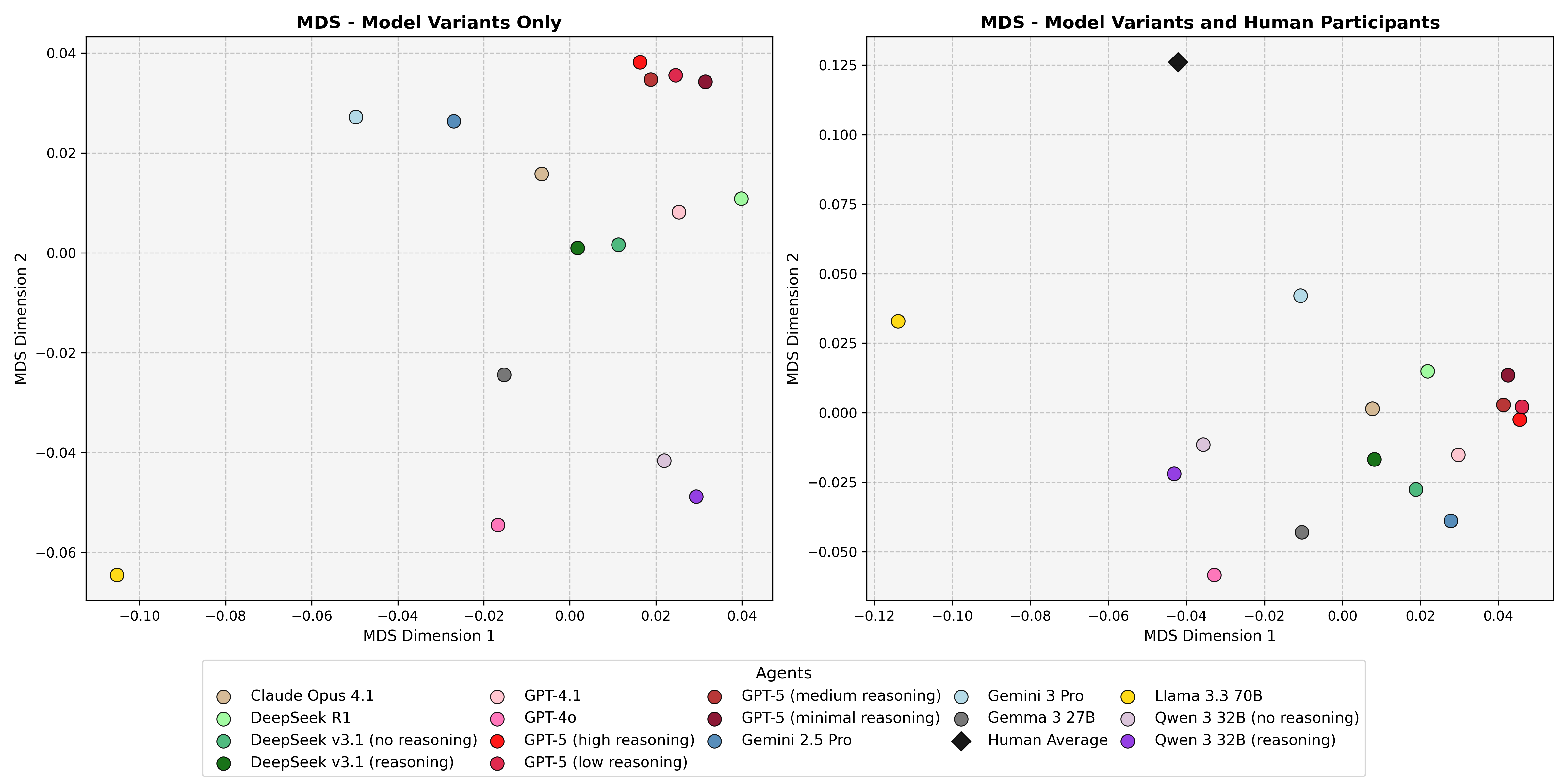} 
    \caption{MDS projections of the semantic embeddings of diagnosis
justifications.}
    \label{fig:justifications}
\end{figure}

The MDS projections of the semantic embeddings of diagnosis justifications are visualized in \autoref{fig:justifications}, with heatmaps for the full matrices with cosine similarity values in \autoref{app:cosines}.

\indent\textbf{Models from a single family (but not necessarily the same series) are generally close semantically.}  This is a trend observable among all the evaluated model families compared to the global model similarity mean of 0.9348: DeepSeek (0.9723, $+$0.0375), GPT (0.9588, $+$0.0240), Gemini (0.9573, $+$0.0225), and Qwen (0.9807, $+$0.0459). Chain-of-thought reasoning had minimal semantic impact, with high similarity between reasoning and non-reasoning variants: 0.9860 for DeepSeek v3.1 Terminus, 0.9807 for Qwen 3 32B, and a between-variant mean of 0.9872 for GPT-5.

\indent\textbf{GPT-4.1 correlates semantically with the GPT-5 cluster, but markedly less so with GPT-4o.} GPT-4.1 
is semantically close to the GPT-5 model variants (0.9687 mean similarity, $+$0.0339 over global model mean), echoing the previous observation. Conversely, GPT-4o is a clear family outlier: while all other within-family pairs span the 0.9573--0.9912 similarity score range, GPT-4o's ties to GPT-4.1 (0.9450) and the four GPT-5 variants (0.9077--0.9111) fall well below this band.

\indent\textbf{Llama 3.3 70B exhibits linguistically atypical outputs, potentially distorting its embeddings.} The distinctiveness of Llama's semantics (mean 0.8648 similarity, $-$0.0700 below global model mean) does not stem from undertaking vastly different diagnostic pathways, 
but from its comparably poor command of the Polish language. Llama's explanations contained incorrect grammar, nonexistent words, or seemingly random ``artifact'' tokens unrelated grammatically and semantically to the content of justifications. This could have appeared as out-of-distribution data to the embedding model, resulting in what is numerically observed as atypical semantics. Interestingly, this did not impede Llama's diagnosis capabilities, as it was, in fact, the ex-aequo third-best performing model (see \autoref{fig:performance_comparison}). \\
\indent\textbf{Several models performing weakly in terms of diagnostic performance also exhibit more atypical semantics.} Three of the four lowest scorers on cumulative diagnostic performance (see \autoref{fig:performance_comparison})---GPT-4o and both Qwen 3 32B variants---are also among the most semantically dissimilar models, with some of the lowest mean similarity to all other models (0.9218, 0.9271, and 0.9330, respectively). Gemma 3 27B, the fourth low scorer, is a partial exception, sitting nearer the cluster center (0.9360). \\
\indent\textbf{Human participant semantics strongly differ from model semantics.} A large observable difference between the general semantic tendencies of human participants compared to models (see \autoref{tab:cosine_means}) can be attributed to differing justification strategies. Human participants provided concise and direct explanations, rarely approaching the 100-word limit. 
Two experts copied excerpts from patient testimonies rather than generating original justifications, while two others produced unusually brief responses: one did not justify healthy-control diagnoses, whereas the other declined to justify low-confidence decisions. Several participants explicitly cut their justifications short, noting insufficient information for a reliable diagnosis. In contrast, all models consistently produced long, detailed, and highly confident justifications, never withholding a diagnosis based on inconclusive data.

\subsubsection{Lexical Features} 
We analyzed the 20 most characteristic features identified for human and model outputs by z-score values (for the full list, see \autoref{app:lexical_features}). The original Polish unigrams were supplemented with their closest English translations and are referred to as such in this section for clarity. 

A higher prevalence of nouns characterized the human-characteristic lexical features group. Specifically, several features point to a human-centric approach to the diagnoses and their subsequent justifications, emphasizing a person's sense of self and their interactions with society: \textit{patient}, \textit{one's own}, \textit{interpersonal}, \textit{others}, \textit{image}. Additionally, a more temporal focus is noticeable through the terms: \textit{time}, \textit{sense of (time)}, \textit{future}. Finally, the observable reluctance to give a definite diagnosis based just on textual data was reflected by mentions of its scarcity by some human participants: \textit{lack of}, \textit{data}.

Conversely, model justifications relied heavily on adjectives, reflecting a strong tendency to categorize and formalize the patients' subjective experiences and symptoms: \textit{persistent}, \textit{severe}, \textit{severity}, \textit{moderate}, \textit{chronic}, \textit{intense}, \textit{rigid}, \textit{entrenched}.
The models also exhibit three pronounced single-feature propensities; heavy focus on seeing patterns in patients' testimonies: \textit{patterns}, comparatively high focus on violent life experiences: \textit{violence}, and a predisposition (contrasting with humans) to create own interpretations of the particulars of the patients' conditions when faced with unsatisfyingly detailed or unclear data: \textit{(I) understand (as)}.

\section{Discussion}
Beyond diagnostic accuracy, this study has several implications for the use of AI technology in clinical mental health contexts. 

Regarding the conceptualization of mental disorders, our results highlight a tension between evolving psychiatric frameworks and the persistence of traditional diagnostic classifications. While human experts performed better on the categorical assessment than the severity assessment task, models achieved relatively balanced scores across both, with a performance preference for categorical diagnosis. This suggests that the current shift toward a dimensional understanding of personality pathology~\cite{APA2022, WHO2022} may be overshadowed by the legacy of earlier categorical systems, whose labels remain prevalent in everyday discourse and online corpora. 

The prominence of such labels in training corpora may also explain the severe underdiagnosing of NPD by models. This could potentially stem from the negative semantic load and social stigma associated with the term ``narcissist'', combined with the current LLM preference alignment training regimen (i.e., RLHF techniques) rewarding agreeable and non-confrontational behavior, sometimes bordering on (social) sycophancy~\cite{Cheng2025, sharma2025understandingsycophancylanguagemodels}. We hypothesize that assigning a stigmatized label to a first-person narrator---essentially ``calling the user a narcissist''---conflicts with these preferences. Importantly, this avoidance does not reflect inability, as models showed higher dimensional recall for NPD cases, indicating sensitivity to severity without using value-laden labels. A similar aversion to stigmatizing terminology may explain the depathologizing bias observed in the GPT models family, where avoidance of psychiatric labels can downplay genuine symptoms and delay necessary treatment~\cite{Semigran2015}. Conversely, BPD is increasingly treated in the literature as a clinical condition rather than a pejorative label. The comparatively extensive body of BPD research relative to other disorders~\cite{Blashfield2000} likely contributes to the models’ inclination to overdiagnose it.

Notably, a critical divergence between humans and LLMs was observed in the assessment of confidence. Models never utilized the ``guessing'' or low-confidence options, whereas human experts frequently expressed uncertainty. This poses a clinical safety risk, as an AI's ability to express doubt in a ``human-like'' manner is a necessary safety feature to ensure users can trust the conveyed uncertainty~\cite{Ulmer2025}.
Paradoxically, although both humans and models demonstrated higher objective performance on the categorical task, they consistently reported higher subjective certainty regarding their dimensional ratings. Future studies could investigate the underlying mechanisms of this metacognitive dissociation, examining why both human and non-human agents perceive their answers regarding dimensional severity as more certain, despite their objective performance superiority in categorical classification.

Furthermore, the semantic analysis reveals that although models may achieve higher accuracy, their ``reasoning'' remains fundamentally distinct from clinical practice. Human justifications were concise, cautious, and patient-centered. Model justifications, conversely, were elaborate, formulaic, and pattern-focused. Additionally, models that showed reduced performance also had more atypical semantics.
This suggests a link between diagnostic accuracy and justification semantics; lower-performing models may have failed to attend to the most clinically informative aspects of the patient narratives. Llama 3.3 70B was a notable exception, achieving high diagnostic accuracy despite generating justifications riddled with linguistic artifacts, nonexistent words, and poor Polish grammar. This can be understood in terms of a multilingual generative-discriminative performance gap: Llama 3.3 70B demonstrates sufficient understanding of Polish to assign accurate diagnoses, but exhibits markedly weaker Polish generation fluency when creating justifications for those decisions. Follow-up research could investigate how the model's performance would change if its diagnostic reasoning were generated before assigning diagnostic scores, rather than as a post-hoc justification.

\section{Conclusion}

In this depth over breadth psychiatric case study, we evaluated state-of-the-art LLMs on raw, first-person patient data, assessed under a novel expert-informed evaluation protocol, and directly compared their performance with human specialists. Our observations suggest that current LLMs demonstrate a surprisingly high level of diagnostic competence, achieving higher diagnostic scores than mental health professionals in the examined context. At the same time, the models demonstrated susceptibility to bias and overconfidence. Taken together, these findings underscore the value of a collaborative human-AI framework, in which the clinical judgment and ethical oversight of experts balance the analytical consistency of models. Such integration has the potential to mitigate the respective limitations of each agent---model bias and human variability---thereby achieving higher diagnostic validity than either could attain alone.

\raggedbottom

\section*{Limitations}
This study has several limitations. First, humans and LLMs were compared on the textual diagnostic modality only, whereas actual psychiatric assessment is inherently multimodal, which could explain the human expert's comparatively lower performance and low inter-rater reliability in this strictly textual format. However, given the rapid post-COVID rise of remote medicine, as well as the more recent surge in AI-driven therapeutic chatbots, the textual modality may gain prominence in the future. 

Second, although a sample size of seven narratives is notably small by the standards of conventional machine learning evaluations, within the context of this study, it reflects a deliberate tradeoff. The limiting factor was the immense cognitive demand of the expert evaluation task. Unlike standard clinical vignettes, the cases used in this study consisted of raw, verbatim transcripts of semi-structured autobiographical interviews drawn from a corpus exceeding 200,000 words, often resulting in over 30 pages of chaotic, non-linear narration per patient. Properly analyzing just one of these narratives required approximately five hours of expert time, as it involved careful reading, cross-referencing temporal inconsistencies, and synthesizing a coherent clinical picture from a ``messy'' life story. Consequently, securing more than 35 hours of active work from practicing, high-level experts proved logistically unfeasible due to time constraints and the significant cognitive toll. Expanding the number of narratives would have introduced a severe risk of expert fatigue; the heavy cognitive load required to review such complex qualitative data at scale would have inevitably compromised the depth and quality of the human diagnostic judgments, thereby undermining the depth over breadth nature of this case study.

Finally, given the complexity of emerging dimensional models of diagnosis comprising not only severity- but also trait-domains-assessment, future research could benefit from also including domain-based between-group comparisons.

\section*{Ethical Considerations}

All procedures performed in this study involving human participants were in accordance with the ethical standards of the institutional and/or national research committee and with the 1964 Helsinki Declaration and its later amendments or comparable ethical standards. Human participants signed an informed consent form by hand. The qualitative interview protocol was approved by the hospital and the Bioethics Committee of Poznan University of Medical Sciences (decision no.  KB-367/23). All personal information was anonymized, ensuring full privacy of the patients' identities.

Mental health diagnosis remains a fundamentally human responsibility requiring empathy, ethical judgment, and contextual understanding beyond mere pattern recognition. Nevertheless, in recent years, patients have already begun to independently use AI-based consumer tools for self-assessment outside established clinical frameworks. We acknowledge the ethical and moral dilemmas associated with the use of AI in clinical research and practice, and, as noted above, advocate that AI technologies be deployed in clinical settings only under the supervision of human experts.

\bibliography{custom}

@book{APA2022,
	title        = {{Diagnostic and statistical manual of mental disorders (5th ed., text rev.)}},
	author       = {{American Psychiatric Association}},
	year         = 2022,
	publisher    = {American Psychiatric Association},
	address      = {Washington, D.C.},
	doi          = {10.1176/appi.books.9780890425787}
}

@article{Ayers2023,
	title        = {{Comparing Physician and Artificial Intelligence Chatbot Responses to Patient Questions Posted to a Public Social Media Forum}},
	author       = {Ayers, John W. and Poliak, Adam and Dredze, Mark and Leas, Eric C. and Zhu, Zechariah and Kelley, Jessica B. and Faix, Dennis J. and Goodman, Aaron M. and Longhurst, Christopher A. and Hogarth, Michael and Smith, Davey M.},
	year         = 2023,
	journal      = {JAMA Internal Medicine},
	publisher    = {American Medical Association},
	volume       = 183,
	number       = 6,
	pages        = {589--596}
}

@article{Barbui2025,
	title        = {{Mental health service coverage and gaps among adults in Europe: a systematic review}},
	author       = {Barbui, Corrado and Alonso, Jordi and Chisholm, Dan and Evans-Lacko, Sara and Keynejad, Roxanne C. and Lazeri, Ledia and Miah, Numan and Valuckiene, Zivile and Gastaldon, Chiara},
	year         = 2025,
	journal      = {The Lancet Regional Health–Europe},
	volume       = 57,
	url          = {https://www.thelancet.com/journals/lanepe/article/PIIS2666-7762(25)00250-9/fulltext}
}

@article{Blashfield2000,
	title        = {{Growth of the literature on the topic of personality disorders}},
	author       = {Blashfield, R. K. and Intoccia, V.},
	year         = 2000,
	journal      = {American Journal of Psychiatry},
	volume       = 157,
	number       = 3,
	pages        = {472--473}
}

@misc{Cheng2025,
      title={{ELEPHANT: Measuring and understanding social sycophancy in LLMs}},
      author={Myra Cheng and Sunny Yu and Cinoo Lee and Pranav Khadpe and Lujain Ibrahim and Dan Jurafsky},
      year={2025},
      eprint={2505.13995},
      archivePrefix={arXiv},
      primaryClass={cs.CL},
      url={https://arxiv.org/abs/2505.13995}, 
}

@article{Cieciuch2022,
	title        = {{Assessment of personality disorder in the ICD-11 diagnostic system: Polish validation of the Personality Inventory for ICD-11}},
	author       = {Cieciuch, Jan and Łakuta, Patryk and Strus, Włodzimierz and Oltmanns, Joshua R. and Widiger, Thomas},
	year         = 2022,
	journal      = {Psychiatria Polska},
	volume       = 56,
	number       = 6,
	pages        = {1185--1202},
	url          = {https://www.psychiatriapolska.pl/Assessment-of-personality-disorder-in-the-ICD-11-diagnostic-system-Polish-validation,138563,0,2.html}
}

@article{diforti2025interrater,
    title        = {{Inter-rater reliability of psychiatric diagnosis: a systematic review and metanalysis}},
    author       = {Di Forti, C. L. and Liccione, D. and Scarpazza, C.},
    year         = 2025,
    journal      = {European Psychiatry},
    volume       = 68,
    number       = {S1},
    pages        = {S191--S192}
}

@article{fiquer2013talking,
    title        = {{Talking bodies: Nonverbal behavior in the assessment of depression severity}},
    author       = {Fiquer, J. T. and Boggio, P. S. and Gorenstein, C.},
    year         = 2013,
    journal      = {Journal of Affective Disorders},
    volume       = 150,
    number       = 3,
    pages        = {1114--1119}
}

@inproceedings{nguyen-etal-2025-large,
	title        = "{Do Large Language Models Align with Core Mental Health Counseling Competencies?}",
	author       = "Nguyen, Viet Cuong  and
      Taher, Mohammad  and
      Hong, Dongwan  and
      Possobom, Vinicius Konkolics  and
      Gopalakrishnan, Vibha Thirunellayi  and
      Raj, Ekta  and
      Li, Zihang  and
      Soled, Heather J.  and
      Birnbaum, Michael L.  and
      Kumar, Srijan  and
      De Choudhury, Munmun",
	year         = 2025,
	booktitle    = "Findings of the Association for Computational Linguistics: NAACL 2025",
	publisher    = "Association for Computational Linguistics",
	address      = "Albuquerque, New Mexico",
	pages        = "7488--7511",
	doi          = "10.18653/v1/2025.findings-naacl.418",
	isbn         = "979-8-89176-195-7",
	url          = "https://aclanthology.org/2025.findings-naacl.418/",
	editor       = "Chiruzzo, Luis  and
      Ritter, Alan  and
      Wang, Lu"
}

@misc{hanafi2025comprehensiveevaluationlargelanguage,
	title        = {{A Comprehensive Evaluation of Large Language Models on Mental Illnesses}},
	author       = {Abdelrahman Hanafi and Mohammed Saad and Noureldin Zahran and Radwa J. Hanafy and Mohammed E. Fouda},
	year         = 2025,
	eprint       = {2409.15687},
	archiveprefix = {arXiv},
	primaryclass = {cs.AI},
	url = {https://arxiv.org/abs/2409.15687}
}

@misc{fouda2025psychiatrybenchmultitaskbenchmarkllms,
	title        = {{PsychiatryBench: A Multi-Task Benchmark for LLMs in Psychiatry}},
	author       = {Aya E. Fouda and Abdelrahamn A. Hassan and Radwa J. Hanafy and Mohammed E. Fouda},
	year         = 2025,
	eprint       = {2509.09711},
	archiveprefix = {arXiv},
	primaryclass = {cs.CL},
	url = {https://arxiv.org/abs/2509.09711}
}

@article{hua2025scoping,
	title        = {{A scoping review of large language models for generative tasks in mental health care}},
	author       = {Hua, Yining and Na, Hongbin and Li, Zehan and Liu, Fenglin and Fang, Xiao and Clifton, David and Torous, John},
	year         = 2025,
	journal      = {npj Digital Medicine},
	publisher    = {Nature Publishing Group},
	volume       = 8,
	number       = 1,
	pages        = 230,
	doi          = {10.1038/s41746-025-01611-4}
}

@article{Harrer2023,
	title        = {{Attention is not all you need: the complicated case of ethically using large language models in healthcare and medicine}},
	author       = {Harrer, Stefan},
	year         = 2023,
	journal      = {eBioMedicine},
	volume       = 90,
	url          = {https://www.thelancet.com/journals/ebiom/article/PIIS2352-3964(23)00077-4/fulltext}
}

@article{Lakuta2023,
	title        = {{Level of Personality Functioning Scale-Brief Form 2.0: Validity and reliability of the Polish adaptation}},
	author       = {Łakuta, Patryk and Cieciuch, Jan and Strus, Włodzimierz and Hutsebaut, Joost},
	year         = 2023,
	journal      = {Psychiatria Polska},
	volume       = 57,
	number       = 2,
	pages        = {247--260},
	url          = {https://www.psychiatriapolska.pl/Level-of-Personality-Functioning-Scale-Brief-Form-2-0-Validity-and-reliability-of,145912,0,2.html}
}

@article{Lawrence2024,
	title        = {{The opportunities and risks of large language models in mental health}},
	author       = {Lawrence, H.R. and Schneider, R.A. and Rubin, S.B. and Matarić, M.J. and McDuff, D.J. and Jones Bell, M.},
	year         = 2024,
	journal      = {JMIR Mental Health},
	volume       = 11,
	url          = {https://mental.jmir.org/2024/1/e54574}
}

@book{McAdams1988,
	title        = {{Power, Intimacy, and the Life Story: Personological Inquiries Into Identity}},
	author       = {McAdams, Dan P.},
	year         = 1988,
	publisher    = {Guilford Press}
}

@article{McBain2025,
	title        = {{Use of Generative AI for Mental Health Advice Among US Adolescents and Young Adults}},
	author       = {McBain, Ryan K. and Bozick, Robert and Diliberti, Melissa and Zhang, Li Ang and Zhang, Fang and Burnett, Alyssa and Kofner, Aaron and Rader, Benjamin and Breslau, Joshua and Stein, Bradley D.},
	year         = 2025,
	journal      = {JAMA Network Open},
	volume       = 8,
	number       = 11,
	pages        = {e2542281},
	url          = {https://jamanetwork.com/journals/jamanetworkopen/article-abstract/2841067}
}

@article{Noda2025,
	title        = {{GPT-4's performance in supporting physician decision-making in nephrology multiple-choice questions}},
	author       = {Noda, Ryunosuke and Tanabe, Kenichiro and Ichikawa, Daisuke and Shibagaki, Yugo},
	year         = 2025,
	journal      = {Scientific Reports},
	volume       = 15,
	number       = 1,
	pages        = 15439,
	url          = {https://www.nature.com/articles/s41598-025-99774-3}
}

@article{Semigran2015,
	title        = {{Evaluation of symptom checkers for self diagnosis and triage: audit study}},
	author       = {Semigran, H. L. and Linder, J. A. and Gidengil, C. and Mehrotra, A.},
	year         = 2015,
	journal      = {BMJ},
	volume       = 351
}

@article{Sun2023,
	title        = {{Low availability, long wait times, and high geographic disparity of psychiatric outpatient care in the US}},
	author       = {Sun, Ching-Fang and Correll, Christoph U. and Trestman, Robert L. and Lin, Yezhe and Xie, Hui and Hankey, Maria Stack and Uymatiao, Raymond Paglinawan and Patel, Riya T. and Metsutnan, Vemmy L. and McDaid, Erin Corinne},
	year         = 2023,
	journal      = {General Hospital Psychiatry},
	volume       = 84,
	pages        = {12--17},
	url          = {https://www.sciencedirect.com/science/article/pii/S0163834323000877}
}

@misc{Ulmer2025,
	title        = {{Anthropomimetic Uncertainty: What Verbalized Uncertainty in Language Models is Missing}},
	author       = {Dennis Ulmer and Alexandra Lorson and Ivan Titov and Christian Hardmeier},
	year         = 2025,
	eprint       = {2507.10587},
	archiveprefix = {arXiv},
	primaryclass = {cs.CL},
	url = {https://arxiv.org/abs/2507.10587}
}

@article{Weekers2023,
	title        = {{Normative data for the LPFS-BF 2.0 derived from the Danish general population and relationship with psychosocial impairment}},
	author       = {Weekers, Laura C. and Sellbom, Martin and Hutsebaut, Joost and Simonsen, Sebastian and Bach, Bo},
	year         = 2023,
	journal      = {Personality and Mental Health},
	volume       = 17,
	number       = 2,
	pages        = {157--164},
	url          = {https://onlinelibrary.wiley.com/doi/abs/10.1002/pmh.1570}
}

@book{WHO2022,
	title        = {{ICD-11: International classification of diseases (11th revision)}},
	author       = {{World Health Organization}},
	year         = 2022,
	publisher    = {World Health Organization},
	url          = {https://icd.who.int/}
}

@article{Workum2025,
	title        = {{Comparative evaluation and performance of large language models on expert level critical care questions: a benchmark study}},
	author       = {Workum, Jessica D. and Volkers, Bas W. S. and van de Sande, Davy and Arora, Sumesh and Goeijenbier, Marco and Gommers, Diederik and van Genderen, Michel E.},
	year         = 2025,
	journal      = {Critical Care},
	volume       = 29,
	pages        = 72,
	url          = {https://pmc.ncbi.nlm.nih.gov/articles/PMC11809097/}
}

@inproceedings{grzybowski-etal-2025-polish,
	title        = "{P}olish-{E}nglish medical knowledge transfer: A new benchmark and results",
	author       = "Grzybowski, {\L}ukasz  and
      Pokrywka, Jakub  and
      Ciesi{\'o}{\l}ka, Micha{\l}  and
      Kaczmarek, Jeremi Ignacy  and
      Kubis, Marek",
	year         = 2025,
	booktitle    = "Findings of the Association for Computational Linguistics: EMNLP 2025",
	publisher    = "Association for Computational Linguistics",
	address      = "Suzhou, China",
	pages        = "9042--9063",
	doi          = "10.18653/v1/2025.findings-emnlp.480",
	isbn         = "979-8-89176-335-7",
	url          = "https://aclanthology.org/2025.findings-emnlp.480/",
	editor       = "Christodoulopoulos, Christos  and
      Chakraborty, Tanmoy  and
      Rose, Carolyn  and
      Peng, Violet"
}

@book{biondi2022clinician,
	title        = {{The Clinician in the Psychiatric Diagnostic Process}},
	year         = 2022,
	publisher    = {Springer Cham},
	address      = {Switzerland},
	doi          = {10.1007/978-3-030-90431-9},
	isbn         = {978-3-030-90430-2},
	editor       = {Biondi, Massimo and Picardi, Angelo and Pallagrosi, Mauro and Fonzi, Laura},
	edition      = {1st}
}

@misc{sharma2025understandingsycophancylanguagemodels,
	title        = {{Towards Understanding Sycophancy in Language Models}},
	author       = {Mrinank Sharma and Meg Tong and Tomasz Korbak and David Duvenaud and Amanda Askell and Samuel R. Bowman and Newton Cheng and Esin Durmus and Zac Hatfield-Dodds and Scott R. Johnston and Shauna Kravec and Timothy Maxwell and Sam McCandlish and Kamal Ndousse and Oliver Rausch and Nicholas Schiefer and Da Yan and Miranda Zhang and Ethan Perez},
	year         = 2025,
	eprint       = {2310.13548},
	archiveprefix = {arXiv},
	primaryclass = {cs.CL},
	url = {https://arxiv.org/abs/2310.13548}
}

@inproceedings{chang-etal-2020-convokit,
	title        = "{{C}onvo{K}it: A Toolkit for the Analysis of Conversations}",
	author       = "Chang, Jonathan P.  and
      Chiam, Caleb  and
      Fu, Liye  and
      Wang, Andrew  and
      Zhang, Justine  and
      Danescu-Niculescu-Mizil, Cristian",
	year         = 2020,
	booktitle    = "Proceedings of the 21th Annual Meeting of the Special Interest Group on Discourse and Dialogue",
	publisher    = "Association for Computational Linguistics",
	address      = "1st virtual meeting",
	pages        = "57--60",
	doi          = "10.18653/v1/2020.sigdial-1.8",
	url          = "https://aclanthology.org/2020.sigdial-1.8/",
	editor       = "Pietquin, Olivier  and
      Muresan, Smaranda  and
      Chen, Vivian  and
      Kennington, Casey  and
      Vandyke, David  and
      Dethlefs, Nina  and
      Inoue, Koji  and
      Ekstedt, Erik  and
      Ultes, Stefan"
}

@article{JMLR:v12:pedregosa11a,
	title        = {{Scikit-learn: Machine Learning in Python}},
	author       = {Fabian Pedregosa and Ga{{\"e}}l Varoquaux and Alexandre Gramfort and Vincent Michel and Bertrand Thirion and Olivier Grisel and Mathieu Blondel and Peter Prettenhofer and Ron Weiss and Vincent Dubourg and Jake Vanderplas and Alexandre Passos and David Cournapeau and Matthieu Brucher and Matthieu Perrot and {{\'E}}douard Duchesnay},
	year         = 2011,
	journal      = {Journal of Machine Learning Research},
	volume       = 12,
	number       = 85,
	pages        = {2825--2830},
	url          = {http://jmlr.org/papers/v12/pedregosa11a.html}
}

@misc{enevoldsen2025mmtebmassivemultilingualtext,
	title        = {{MMTEB: Massive Multilingual Text Embedding Benchmark}},
	author       = {Kenneth Enevoldsen and Isaac Chung and Imene Kerboua and Márton Kardos and Ashwin Mathur and David Stap and Jay Gala and Wissam Siblini and Dominik Krzemiński and Genta Indra Winata and Saba Sturua and Saiteja Utpala and Mathieu Ciancone and Marion Schaeffer and Gabriel Sequeira and Diganta Misra and Shreeya Dhakal and Jonathan Rystrøm and Roman Solomatin and Ömer Çağatan and Akash Kundu and Martin Bernstorff and Shitao Xiao and Akshita Sukhlecha and Bhavish Pahwa and Rafał Poświata and Kranthi Kiran GV and Shawon Ashraf and Daniel Auras and Björn Plüster and Jan Philipp Harries and Loïc Magne and Isabelle Mohr and Mariya Hendriksen and Dawei Zhu and Hippolyte Gisserot-Boukhlef and Tom Aarsen and Jan Kostkan and Konrad Wojtasik and Taemin Lee and Marek Šuppa and Crystina Zhang and Roberta Rocca and Mohammed Hamdy and Andrianos Michail and John Yang and Manuel Faysse and Aleksei Vatolin and Nandan Thakur and Manan Dey and Dipam Vasani and Pranjal Chitale and Simone Tedeschi and Nguyen Tai and Artem Snegirev and Michael Günther and Mengzhou Xia and Weijia Shi and Xing Han Lù and Jordan Clive and Gayatri Krishnakumar and Anna Maksimova and Silvan Wehrli and Maria Tikhonova and Henil Panchal and Aleksandr Abramov and Malte Ostendorff and Zheng Liu and Simon Clematide and Lester James Miranda and Alena Fenogenova and Guangyu Song and Ruqiya Bin Safi and Wen-Ding Li and Alessia Borghini and Federico Cassano and Hongjin Su and Jimmy Lin and Howard Yen and Lasse Hansen and Sara Hooker and Chenghao Xiao and Vaibhav Adlakha and Orion Weller and Siva Reddy and Niklas Muennighoff},
	year         = 2025,
	eprint       = {2502.13595},
	archiveprefix = {arXiv},
	primaryclass = {cs.CL},
	url = {https://arxiv.org/abs/2502.13595}
}

@misc{comanici2025gemini25pushingfrontier,
	title        = {{Gemini 2.5: Pushing the Frontier with Advanced Reasoning, Multimodality, Long Context, and Next Generation Agentic Capabilities}},
	author       = {{Gemini Team}},
	year         = 2025,
	eprint       = {2507.06261},
	archiveprefix = {arXiv},
	primaryclass = {cs.CL},
	url = {https://arxiv.org/abs/2507.06261}
}

@misc{openai2024gpt4ocard,
	title        = {{GPT-4o System Card}},
	author       = {OpenAI},
	year         = 2024,
	eprint       = {2410.21276},
	archiveprefix = {arXiv},
	primaryclass = {cs.CL},
	url = {https://arxiv.org/abs/2410.21276}
}

@misc{deepseekai2025deepseekr1incentivizingreasoningcapability,
	title        = {{DeepSeek-R1: Incentivizing Reasoning Capability in LLMs via Reinforcement Learning}},
	author       = {DeepSeek-AI},
	year         = 2025,
	eprint       = {2501.12948},
	archiveprefix = {arXiv},
	primaryclass = {cs.CL},
	url = {https://arxiv.org/abs/2501.12948}
}

@misc{grattafiori2024llama3herdmodels,
	title        = {{The Llama 3 Herd of Models}},
	author       = {{Llama Team}},
	year         = 2024,
	eprint       = {2407.21783},
	archiveprefix = {arXiv},
	primaryclass = {cs.AI},
	url = {https://arxiv.org/abs/2407.21783}
}

@misc{gemmateam2025gemma3technicalreport,
	title        = {{Gemma 3 Technical Report}},
	author       = {{Gemma Team}},
	year         = 2025,
	eprint       = {2503.19786},
	archiveprefix = {arXiv},
	primaryclass = {cs.CL},
	url ={https://arxiv.org/abs/2503.19786}
}

@misc{yang2025qwen3technicalreport,
	title        = {{Qwen3 Technical Report}},
	author       = {{Qwen Team}},
	year         = 2025,
	eprint       = {2505.09388},
	archiveprefix = {arXiv},
	primaryclass = {cs.CL},
	url = {https://arxiv.org/abs/2505.09388}
}

@misc{deepseekai2025deepseekv3technicalreport,
	title        = {{DeepSeek-V3 Technical Report}},
	author       = {DeepSeek-AI},
	year         = 2025,
	eprint       = {2412.19437},
	archiveprefix = {arXiv},
	primaryclass = {cs.CL},
	url = {https://arxiv.org/abs/2412.19437}
}

@article{kruskal1964multidimensional,
	title        = {{Multidimensional scaling by optimizing goodness of fit to a nonmetric hypothesis}},
	author       = {Kruskal, Joseph B.},
	year         = 1964,
	journal      = {Psychometrika},
	volume       = 29,
	number       = 1,
	pages        = {1--27},
	doi          = {10.1007/BF02289565}
}

@article{shi2025mentalqlm,
	title        = {{{MentalQLM}: A Lightweight Large Language Model for Mental Healthcare Based on Instruction Tuning and Dual {LoRA} Modules}},
	author       = {Shi, Jiayu and Wang, Zexiao and Zhou, Jiandong and Liu, Chengyu and Sun, Poly Z. H. and Zhao, Erying and Lu, Lei},
	year         = 2025,
	journal      = {IEEE Journal of Biomedical and Health Informatics},
	volume       = {Early Access},
	pages        = {1--12},
	doi          = {10.1109/JBHI.2025.3594133}
}

@article{monroe2008fightin,
	title        = {{Fightin' Words: Lexical Feature Selection and Evaluation for Identifying the Content of Political Conflict}},
	author       = {Monroe, Burt L. and Colaresi, Michael P. and Quinn, Kevin M.},
	year         = 2008,
	journal      = {Political Analysis},
	volume       = 16,
	number       = 4,
	pages        = {372--403},
	doi          = {10.1093/pan/mpn018}
}

@misc{google2025gemini3,
	title        = {{Gemini 3 Pro Model Card}},
	author       = {{Google}},
	year         = 2025,
	note         = {Accessed: 2025-12-18},
	howpublished = {\url{https://storage.googleapis.com/deepmind-media/Model-Cards/Gemini-3-Pro-Model-Card.pdf}}
}

@misc{anthropic2025claude4,
	title        = {{System Card: Claude Opus 4 \& Claude Sonnet 4}},
	author       = {{Anthropic}},
	year         = 2025,
	note         = {Accessed: 2025-12-18},
	howpublished = {\url{https://www-cdn.anthropic.com/4263b940cabb546aa0e3283f35b686f4f3b2ff47.pdf}}
}

@misc{openai2025gpt41,
	title        = {{Introducing GPT-4.1 in the API}},
	author       = {{OpenAI}},
	year         = 2025,
	note         = {Accessed: 2025-12-18},
	howpublished = {\url{https://openai.com/index/gpt-4-1/}}
}

@misc{openai2025gpt5,
	title        = {{GPT-5 System Card}},
	author       = {{OpenAI}},
	year         = 2025,
	note         = {Accessed: 2025-12-18},
	howpublished = {\url{https://cdn.openai.com/gpt-5-system-card.pdf}}
}
\onecolumn
\appendix

\setcounter{figure}{0}
\renewcommand{\thefigure}{\thesection.\arabic{figure}}

\setcounter{table}{0}
\renewcommand{\thetable}{\thesection.\arabic{table}}

\section{Clinical Assessment Instruments}
\label{app:instruments}
Levels of personality functioning and maladaptive trait domains were assessed using the Level of Personality Functioning---Brief Scale 2.0 (LPFS-BF 2.0)~\cite{Lakuta2023} and the Personality Inventory for ICD-11~\cite{Cieciuch2022}, ensuring comparability across individuals. LPFS-BF 2.0 scores were interpreted relative to established cut-offs~\cite{Weekers2023}, yielding a dimensional classification from subclinical to extreme impairment (0--48), independent of the categorical BPD/NPD labels.

\section{Prompts}
\label{app:model_prompt}

\medskip

\begin{tcolorbox}[breakable, colback=blue!10, colframe=blue!40, boxrule=0.5mm, title=Original Polish prompt used in the study, coltitle=black, halign title=center]
    
Jesteś doświadczonym specjalistą zdrowia psychicznego. Twoim zadaniem jest przeprowadzenie wstępnej oceny diagnostycznej na podstawie przedstawionego poniżej tekstu autobiograficznego. Wykonaj 6-etapową analizę, prezentując swoje wnioski w ściśle określonym formacie. \\

\# OCENA DIAGNOSTYCZNA \\

\#\# Krok 1 \\

Przypisz diagnozę kategorialną zaburzenia osobowości. Pamiętaj, że możliwa jest opcja ``Brak zaburzenia''. \\

\#\# Krok 2 \\

Oceń stopień pewności co do poprawności swojej odpowiedzi z Kroku 1, używając skali 1--4, gdzie: 1 = zgadywałem/am, 2 = trochę pewny/a, 3 = dość pewny/a, 4 = całkowicie pewny/a. \\

\#\# Krok 3 \\

Uzasadnij swoją decyzję diagnostyczną z Kroku 1. Wskaż kluczowe fragmenty danych, które popierają twoją ocenę. Następnie, odnieś się do tego czym są dla ciebie zaburzenia osobowości i wyjaśnij, jak rozumiesz ich źródła. Limit uzasadnienia to 100 słów. \\

\#\# Krok 4 \\

Oceń stopień nasilenia zaburzenia osobowości, używając skali 0--3, gdzie: 0 = brak, 1 = łagodne, 2 = umiarkowane, 3 = ciężkie. \\

\#\# Krok 5 \\

Oceń stopień pewności co do poprawności oceny nasilenia z Kroku 4, używając skali 1--4, gdzie: 1 = zgadywałem/am, 2 = trochę pewny/a, 3 = dość pewny/a, 4 = całkowicie pewny/a. \\

\#\# Krok 6 \\

Uzasadnij swoją odpowiedź z Kroku 4. Wskaż kluczowe fragmenty danych, które popierają Twoją ocenę. Następnie, odnieś się do tego czym są dla ciebie zaburzenia osobowości i wyjaśnij, jak rozumiesz ich źródła. Limit uzasadnienia to 100 słów.\\
Udziel odpowiedzi, używając poniższego szablonu.\\
Analiza Przypadku [Numer przypadku]\\
Diagnoza Kategorialna: [Twoja odpowiedź]\\
Pewność Diagnozy (1--4): [Twoja odpowiedź]\\
Uzasadnienie Diagnozy (do 100 słów): [Twoja odpowiedź]\\
Ocena Nasilenia (0--3): [Twoja odpowiedź]\\
Pewność Oceny Nasilenia (1--4): [Twoja odpowiedź]\\
Uzasadnienie Oceny Nasilenia (do 100 słów): [Twoja odpowiedź]\\
Teraz zapoznaj się z tekstem, przeprowadź analizę i odpowiedz zgodnie z podanym formatem. \\

\#\# TEKST AUTOBIOGRAFICZNY \\

\{text\} \\

\#\# ODPOWIEDŹ

\end{tcolorbox}

\smallskip

\begin{tcolorbox}[breakable, colback=red!10, colframe=red!40, boxrule=0.5mm, title=English translation of the original prompt, coltitle=black, halign title=center]
You are an experienced mental health professional. Your task is to conduct a preliminary diagnostic assessment based on the autobiographical text presented below. Perform a 6-step analysis, presenting your conclusions in a strictly defined format. \\
 \\
\# DIAGNOSTIC ASSESSMENT \\
 \\
\#\# Step 1 \\ \\
Assign a categorical personality disorder diagnosis. Remember that the option ``No disorder'' is possible. \\
 \\
\#\# Step 2 \\ \\
Rate the degree of certainty regarding the correctness of your answer from Step 1, using a scale of 1--4, where: 1 = I was guessing, 2 = somewhat certain, 3 = fairly certain, 4 = completely certain. \\
 \\
\#\# Step 3 \\ \\
Justify your diagnostic decision from Step 1. Indicate key data fragments that support your assessment. Then, refer to what personality disorders are to you and explain how you understand their origins. The justification limit is 100 words. \\
 \\
\#\# Step 4 \\ \\
Rate the severity of the personality disorder, using a scale of 0--3, where: 0 = none, 1 = mild, 2 = moderate, 3 = severe. \\
 \\
\#\# Step 5 \\ \\
Rate the degree of certainty regarding the correctness of the severity assessment from Step 4, using a scale of 1--4, where: 1 = I was guessing, 2 = somewhat certain, 3 = fairly certain, 4 = completely certain. \\
 \\
\#\# Step 6 \\ \\
Justify your answer from Step 4. Indicate key data fragments that support your assessment. Then, refer to what personality disorders are to you and explain how you understand their origins. The justification limit is 100 words. \\
Provide your answer using the template below. \\
Case Analysis [Case Number] \\
Categorical Diagnosis: [Your answer] \\
Diagnosis Certainty (1--4): [Your answer] \\
Diagnosis Justification (up to 100 words): [Your answer] \\
Severity Assessment (0--3): [Your answer] \\
Severity Assessment Certainty (1--4): [Your answer] \\
Severity Assessment Justification (up to 100 words): [Your answer]

Now familiarize yourself with the text, perform the analysis, and answer according to the provided format. \\

\#\# AUTOBIOGRAPHICAL TEXT \\

\{text\} \\

\#\# ANSWER

\end{tcolorbox}

\section{Diagnostic Labels}
\label{sec:appendix_table}

\begin{table}[ht]
\centering
\small 
\begin{tabular}{lrr}
\toprule
\textbf{Label} & \textbf{Count} & \textbf{Percentage} \\
\midrule
BPD & 204 & 53.97\% \\
HC (Healthy) & 81 & 21.43\% \\
AvPD & 57 & 15.08\% \\
NPD & 10 & 2.65\% \\
Unspecified PD & 6 & 1.59\% \\
DPD & 5 & 1.32\% \\
AD/UNS & 4 & 1.06\% \\
Cannot diagnose & 3 & 0.79\% \\
DePD & 2 & 0.53\% \\
ASPD & 2 & 0.53\% \\
StPD & 1 & 0.26\% \\
OPD & 1 & 0.26\% \\
MPD & 1 & 0.26\% \\
HPD & 1 & 0.26\% \\
\bottomrule
\end{tabular}
\captionsetup{justification=centering, singlelinecheck=false}
\caption{Distribution of diagnostic labels.}
\label{tab:diagnoses_distribution}
\end{table}

\noindent The diagnostic labels presented in \autoref{tab:diagnoses_distribution} correspond to the following clinical categories: BPD: Borderline Personality Disorder; NPD: Narcissistic Personality Disorder; AvPD: Avoidant Personality Disorder; HC: Healthy; StPD: Schizotypal Personality Disorder; DPD: Dependent Personality Disorder; DePD: Depressive Personality Disorder; AD/UNS: Anxiety Disorder/Unspecified; ASPD: Antisocial Personality Disorder; OPD: Obsessive Personality Disorder; MPD: Masochistic Personality Disorder; HPD: Histrionic Personality Disorder.

\clearpage

\section{Diagnostic Performance Heatmap}

\label{sec:appendix_d}
\begin{figure}[H]
        \captionsetup{justification=centering, singlelinecheck=false}
        \includegraphics[width=\columnwidth]{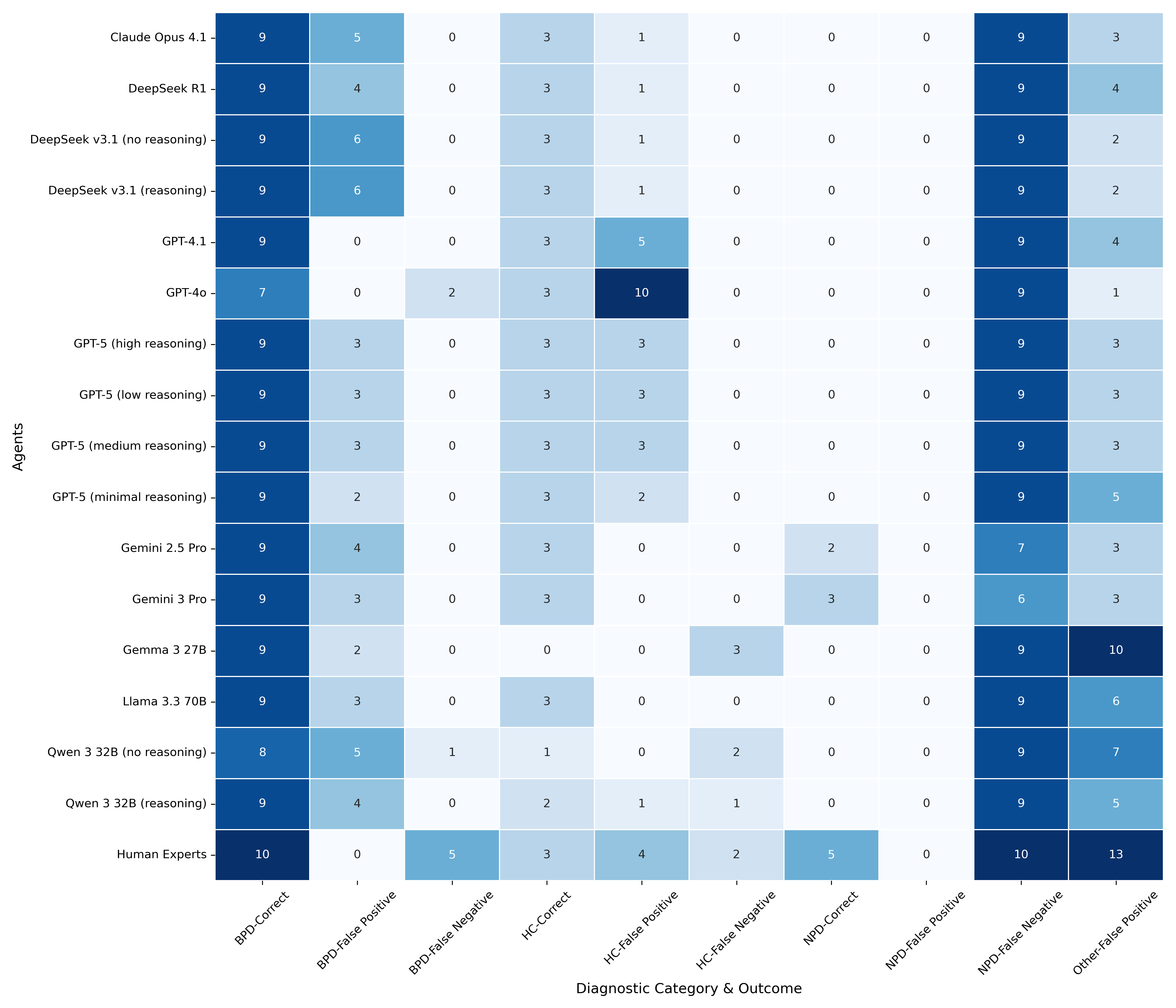}\
        \caption{Diagnostic performance heatmap of human experts and models.}
        \label{fig:heatmap}
\end{figure}

The ``Human Experts'' row aggregates the performance of $N=5$ professionals; the sixth professional declined to provide definitive categorical labels, thus their data was included only in the dimensional evaluation. A False Positive indicates diagnosing a disorder absent in the ground truth, while a False Negative denotes failing to identify a present condition. Each model row reflects $N=21$ trials (7 cases x 3 runs), whereas the human row corresponds to $N=35$ independent assessments (7 cases x 5 experts).

\section{Semantic Justification Embedding Creation Process}
\label{app:justification_process}
The process of mapping the diagnosis justifications obtained through our procedure onto a high-dimensional semantic embedding space comprised four steps. 

\begin{enumerate}
    \item The texts of all justifications were aggregated into groupings. For humans, due to the small amount of data relative to model outputs, a single grouping was created from all justifications for a single human participant. For each model, we created two separate groupings: one based on all categorical justifications and the other based on all dimensional justifications.
    \item Before the embedding process, simple text pre-processing with regular expressions was applied to remove LLM-characteristic text formatting artifacts, such as Markdown syntax and redundant whitespace.
    \item  Each grouping was converted into a dense embedding using the chosen \texttt{BAAI/bge-multilingual-gemma2} embedding model. The model operated in 16-bit precision on a single NVIDIA A100 (40GB) GPU, with a batch size of 8 and default remaining hyperparameter values.
    \item  A summary embedding representing the semantic contents of justifications was created for each model by first calculating the mean value for the categorical and dimensional grouping embeddings separately, and subsequently averaging these two values. For human participants, because there was no categorical-dimensional grouping, a single summary embedding was derived by averaging the embeddings of all individuals.
\end{enumerate}

\section{Justification Embedding Pairwise Cosine Similarity Heatmap}
\label{app:cosines}

\setcounter{figure}{0}

\begin{figure}[H]
    \centering
    \includegraphics[width=0.9\columnwidth]{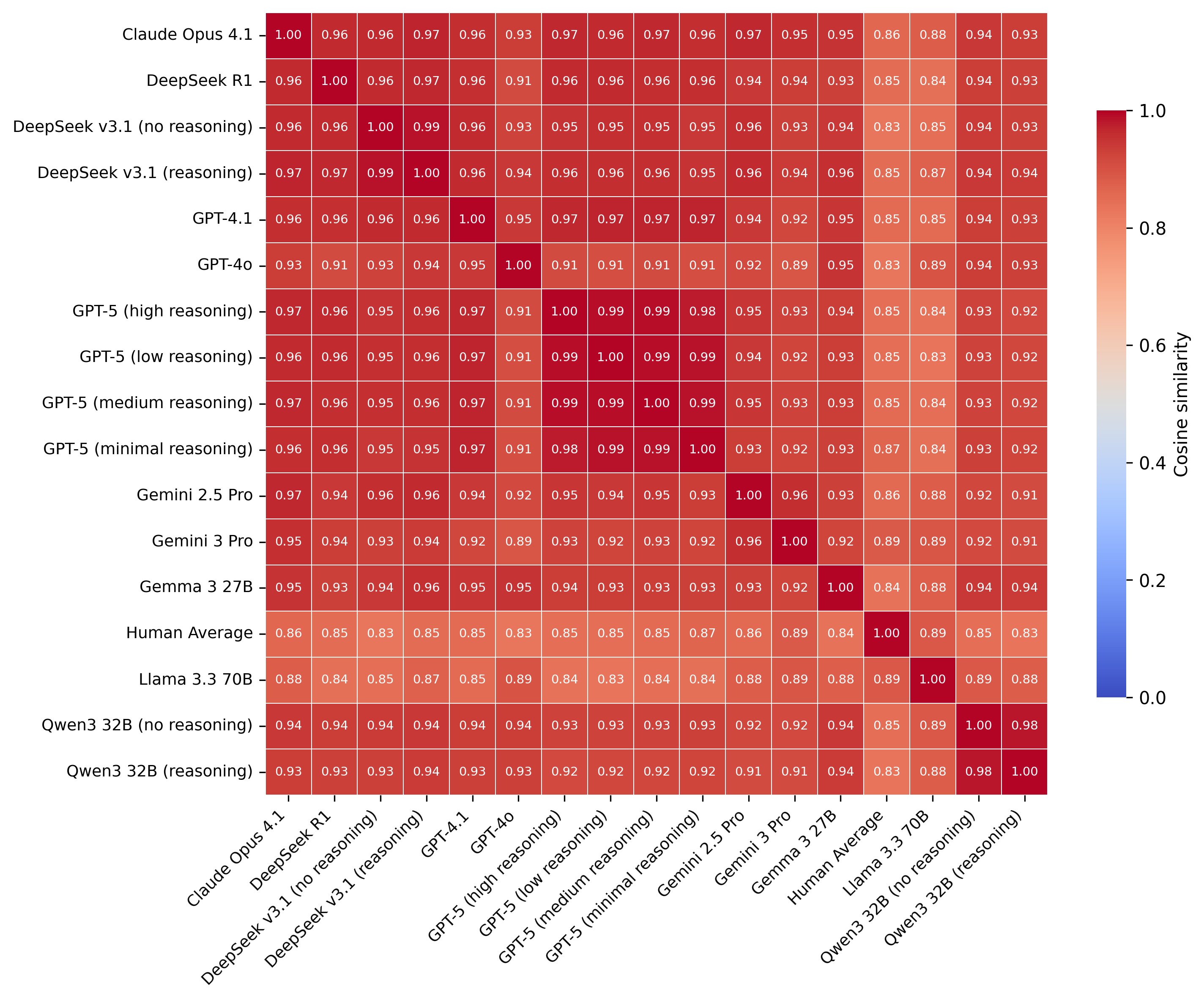}
    \label{fig:cosine_severity}
    \caption{Pairwise cosine similarities for the diagnosis justification embeddings.}
\end{figure}

\setcounter{table}{0}

\begin{table}[h]
\centering
\begin{tabular}{lr}
\hline
\textbf{Measure} & \textbf{Mean} \\
\hline
Models (pairwise) & 0.9348 $\pm$ 0.0338 \\
Human Average vs. Models (pairwise) & 0.8524 $\pm$ 0.0162 \\
\hline
Gap & $+$0.0824 \\
\hline
\end{tabular}
\caption{Cosine similarity means for the diagnosis justification embeddings.}
\label{tab:cosine_means}
\end{table}

\clearpage

\section{Dimensionality Reduction Hyperparameters}
\label{app:hyperparameters}

\setcounter{table}{0}

\begin{table}[h]
    \centering
    \begin{tabular}{lrr}
        \hline
        \textbf{Algorithm} & \textbf{Parameter} & \textbf{Value} \\
        \hline
        MDS & \texttt{dissimilarity} & precomputed \\
            & \texttt{n\_components} & 2 \\
            & \texttt{random\_state} & 42 \\
        \hline
    \end{tabular}
    \caption{Dimensionality reduction algorithm hyperparameter values.}
    \label{tab:hyperparams}
\end{table}

\noindent Hyperparameters not present in the table retained the default values set in the algorithm implementation.

\section{Most Representative Lexical Features}
\label{app:lexical_features}

\setcounter{figure}{0}

\begin{figure}[H]
    \centering
    \includegraphics[width=\columnwidth]{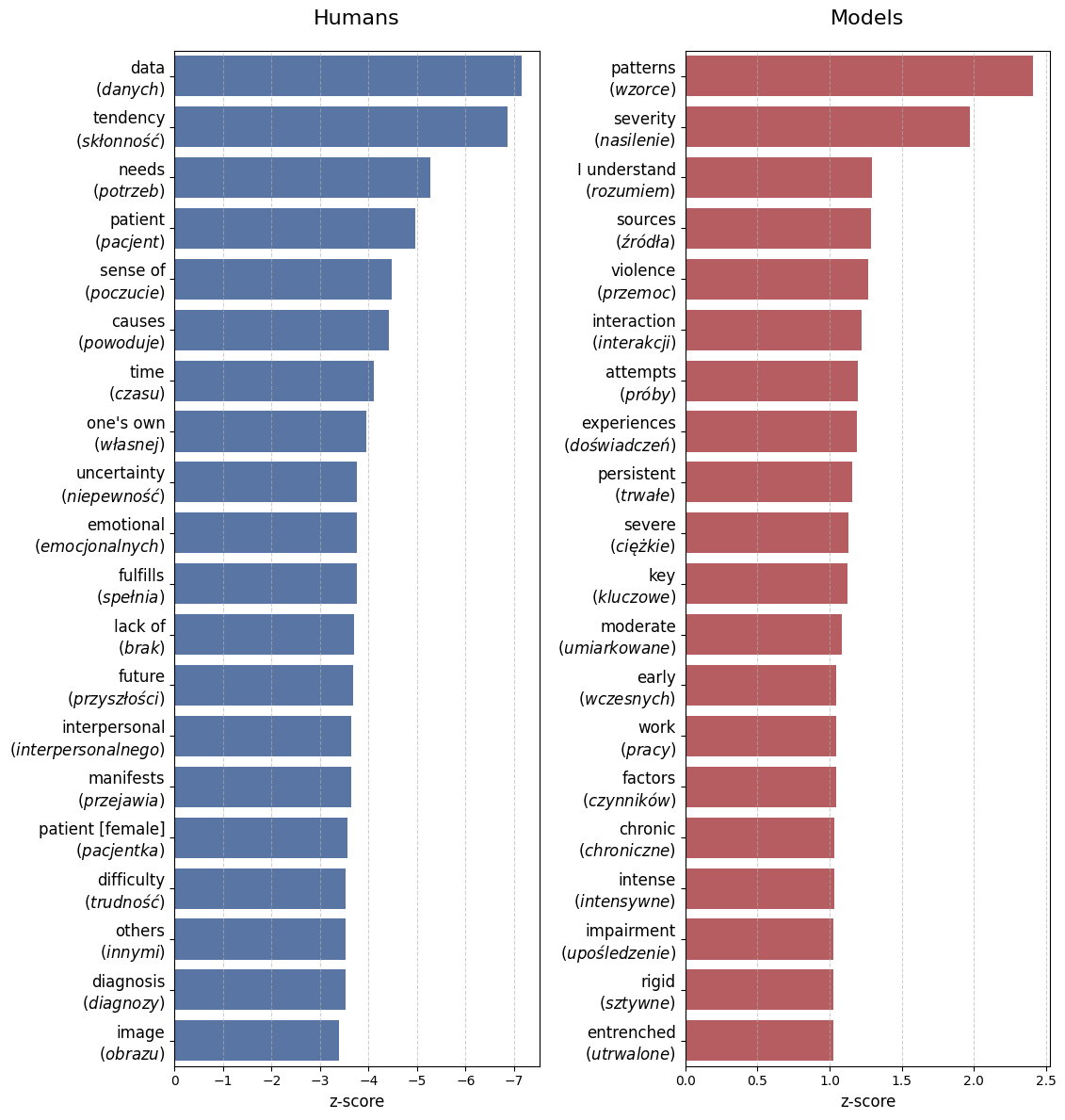}
    \caption{The 20 most representative lexical features for human experts and models by z-score.}
    \label{fig:ngrams}
\end{figure}

\end{document}